\documentclass[conference]{IEEEtran}
\IEEEoverridecommandlockouts

\usepackage{xfp}

\usepackage{cite}
\usepackage{amsmath,amssymb,amsfonts}
\usepackage{algorithmic}
\usepackage{graphicx}
\usepackage{textcomp}
\usepackage{xcolor}
\usepackage{multirow}
\def\BibTeX{{\rm B\kern-.05em{\sc i\kern-.025em b}\kern-.08em
    T\kern-.1667em\lower.7ex\hbox{E}\kern-.125emX}}
\begin{document}

\title{Densification and forecasting of Sentinel-2 time series from multimodal SAR and Optical satellite data using deep generative models}

\author{\IEEEauthorblockN{Véronique Defonte}
\IEEEauthorblockA{\textit{Thales Services Numériques }\\
31670 Labège, France }
\and
\IEEEauthorblockN{Dawa Derksen}
\IEEEauthorblockA{\textit{Centre National d'Etudes Spatiales} \\
31400 Toulouse, France}
\and
\IEEEauthorblockN{Alexandre Constantin}
\IEEEauthorblockA{\textit{Centre National d'Etudes Spatiales} \\
31400 Toulouse, France}
\and
\IEEEauthorblockN{Bastien Nespoulous}
\IEEEauthorblockA{\textit{Thales Services Numériques }\\
31670 Labège, France }
}

\maketitle

\begin{abstract}
Optical satellite image time series are extensively used in many Earth observation applications, including agriculture, climate monitoring, and land surface analysis. However, clouds and swath edges result in irregular sampling along the temporal dimension, limiting continuous monitoring. To address this issue, a growing body of work has focused on temporal densification and reconstruction of satellite image time series, with the objective of filling missing or cloud-contaminated observations within the temporal extent of the available data. While these approaches improve temporal continuity, they are inherently restricted to the reconstruction of the gaps within the observed time periods, and do not address the prediction of future observations. This work proposes a probabilistic deep learning framework for the densification and forecasting of Sentinel-2 time series by generating optical images at arbitrary past or future dates. The approach leverages multimodal satellite data by jointly exploiting Sentinel-2 optical and Sentinel-1 SAR observations. Unlike most existing works, we propose to focus on the uncertainty of the generated images. Experimental results demonstrate effective densification and forecasting, on sparse and temporally misaligned time series.
\end{abstract}


\section{Introduction}

Optical satellite image time series have become a key component of Earth observation, supporting a wide range of applications such as land surface monitoring \cite{land_cover_maps}, agriculture \cite{world_cereal}, ecosystem dynamics \cite{ecosystems}, and climate-related studies \cite{flood_mapping}. Thanks to their rich spectral content, optical observations enable the characterization of vegetation, soil, and surface properties through a variety of spectral indices and biophysical variables. The increasing availability of long-term satellite observations has enabled the analysis of temporal patterns, seasonal dynamics, and abrupt changes at fine spatial scales. However, despite these advances, the practical use of optical satellite time series remains strongly limited by acquisition gaps and irregular sampling, primarily caused by cloud cover and varying observation conditions. A study over 12 years of optical data acquired by the MODIS sensor concluded that, on average, clouds occlude $67\%$ of the Earth’s surface and $55\%$ of the land surface at any point in time \cite{modis_cloud}.

To address these limitations, recent works have increasingly explored the fusion of optical and SAR imagery using deep learning for optical image reconstruction, motivated by the need for continuous, cloud-free Earth surface monitoring, commonly studied under the umbrella of cloud removal. Since SAR data, such as Sentinel-1, are not affected by cloud cover, they provide essential complementary information for this task.

A first line of work addresses single-date image reconstruction, where a cloud-free Sentinel-2 image is generated using observations acquired on that date, often in combination with co-registered Sentinel-1 SAR data. Methods such as DSen2-CR \cite{DSen2_CR} formulate cloud removal as a SAR-guided reconstruction problem using cycle-consistent GANs, where co-registered Sentinel-1 SAR and Sentinel-2 optical observations are concatenated and jointly processed to reconstruct cloud-free optical images. Building on recent advances in attention mechanisms inspired by Transformer architectures, the GLF-CR framework \cite{GLF_CR} introduces global–local attention to selectively inject SAR-derived structural information into the optical representation. More recently, \cite{GLCdiffcr} introduced GLCdiffcr, a diffusion-based framework incorporating global–local interactions and conditional consistency constraints for SAR-guided optical image cloud removal. Although effective, these approaches operate strictly at the image level and do not exploit temporal information.

To reduce the ambiguity inherent to single-date reconstruction, a distinct line of work explores sequence-to-image generation, where a multi-temporal sequence of satellite images is used as input to generate a single cloud-free optical image. These approaches exploit temporal redundancy to aggregate complementary information across multiple observations, but differ from sequence-to-sequence models in that temporal information is compressed into a single output representation. \cite{stgan} propose a spatiotemporal generator networks (STGAN), where multiple cloudy Sentinel-2 images acquired at different dates are jointly processed through branched encoder–decoder architectures. Temporal information is handled by extracting features independently from each input image and progressively aggregating them through feature concatenation and shared decoding layers, enabling a many-to-one mapping that synthesizes a single cloud-free image. Building upon this formulation, the SEN12MS-CR-TS benchmark \cite{SEN12MS} adopts a similar spatiotemporal generator architecture while extending it to a multimodal setting, where co-registered Sentinel-1 SAR images are concatenated with the corresponding Sentinel-2 observations at each date.

Moving from many-to-one aggregation toward explicit temporal modeling, a growing body of work formulates Sentinel-2 reconstruction as a sequence-to-sequence problem, focusing on the reconstruction of complete and gap-free Sentinel-2 time series from irregular and cloud-contaminated observations acquired by a single optical sensor. In this mono-sensor, optical-only setting, \cite{U-TILISE} adopts a sequence-to-sequence architecture that combines a convolutional U-Net for spatial–spectral modeling with temporal self-attention to capture long-range dependencies across time. GANFilling \cite{GANFilling} further introduces a generative sequence-to-sequence framework based on GANs with convolutional LSTM layers, explicitly modeling spatio-temporal dynamics and demonstrating robust gap filling of Sentinel-2 reflectance series, as well as benefits for downstream forecasting tasks. However, optical-only sequence models may struggle under prolonged cloud cover or highly irregular sampling, where optical observations alone provide insufficient information.

The most recent approaches address these limitations by performing multimodal time-series generation, jointly modeling temporal dynamics and cross-sensor complementarity. The RESTORE-DiT framework \cite{RESTORE-DiT} proposes a multimodal diffusion-based framework for reconstructing cloud-contaminated Sentinel-2 images. Sentinel-1 SAR observations are integrated into the diffusion process through multimodal cross-attention, guiding the optical reconstruction in a temporally consistent manner and enabling robust generation under persistent cloud cover and irregular acquisition patterns. 
In a complementary direction, the TAMRF model \cite{Michel2026} addresses the fusion of Satellite Image Time Series (SITS) acquired from multiple optical sensors with different spatial resolutions and temporal sampling, such as Sentinel-2 and Landsat-8/9. Given multi-resolution and irregularly sampled time series over the same area, TAMRF learns a latent representation from which it can reconstruct observations at arbitrary query dates, at the highest available spatial resolution, and free of clouds.

Despite their effectiveness for cloud removal and gap filling, current reconstruction-based methods remain fundamentally limited to interpolation within the temporal range of available observations. As a result, they are not designed to generate plausible optical images at arbitrary future dates, nor to explicitly quantify the uncertainty associated with their predictions. Yet, the ability to extrapolate beyond observed time points and to provide uncertainty estimates is critical for many downstream applications, including forecasting, scenario analysis, and decision support, where understanding model confidence is as important as the prediction itself.

In this work, we propose a probabilistic framework for producing optical satellite images at a specified target date from irregular and sparse multimodal time series, explicitly designed to preserve fine-grained spatial structure while modeling long-range temporal dependencies. By formulating the problem as target-conditioned image generation, the proposed approach naturally supports both temporal interpolation and extrapolation. Moreover, the model explicitly predicts pixel-wise uncertainty alongside the generated optical image, providing valuable information about prediction reliability. Unlike many existing methods, our approach does not rely on explicit cloud masks or temporally aligned multimodal inputs, and instead learns to selectively leverage available observations in a fully end-to-end manner.

\section{Method}

We consider sparse and irregular multimodal satellite time series composed of Sentinel-2 optical and Sentinel-1 RADAR acquisitions. For each sample, Sentinel-2 data are represented as a sequence of optical patches $\mathbf{X}^{S2} \in \mathbb{R}^{T_{S2} \times C_{S2} \times H \times W}$, where $T_{S2}$ denotes the number of available optical acquisitions, $C_{S2}$ the RGB-NIR bands, and $H \times W$ the spatial resolution. Similarly, Sentinel-1 data are represented as $\mathbf{X}^{S1} \in \mathbb{R}^{T_{S1} \times C_{S1} \times H \times W}$, with $T_{S1}$ the number of radar acquisitions and $C_{S1}$ corresponding to the VV and VH polarizations.

Given a target date $d_{target}$, the objective is to generate a complete Sentinel-2 optical patch $\hat{\mathbf{Y}}(d_{target}) \in \mathbb{R}^{C_{S2} \times H \times W}$ along with an associated uncertainty map, conditioned on the target date and the available multimodal observations with their acquisition times. This task is formulated as a conditional image generation problem, where the model must infer spatially coherent optical reflectance values at the target date despite temporal sparsity and missing data.

\subsection{Overview}

\begin{figure*}[htbp]
\centering
\includegraphics[width=0.7\textwidth]{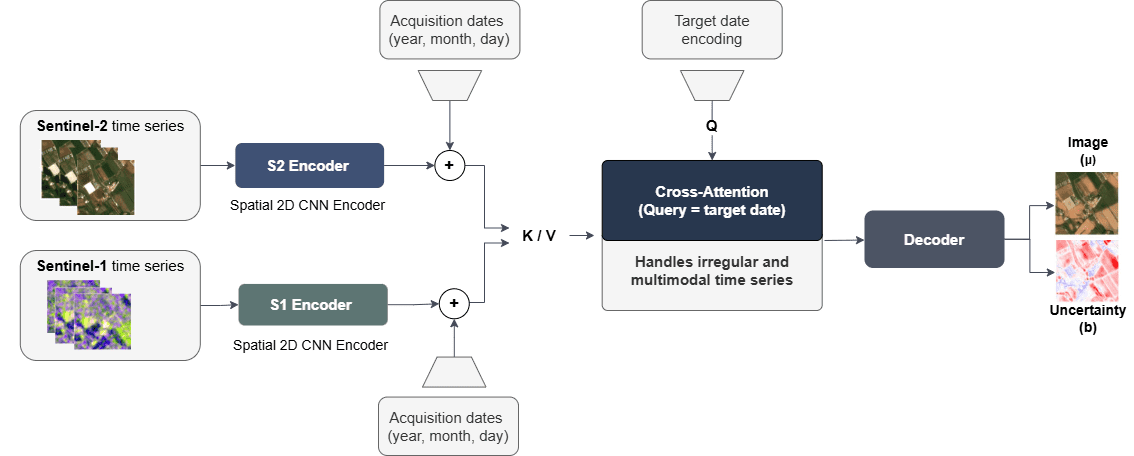}
\caption{Overview of the proposed model for generating an optical image at a target date from sparse and irregular Sentinel-1 and Sentinel-2 time series. Spatial features are first extracted independently for each modality using 2D convolutional encoders with spatial pyramid pooling. Acquisition dates are encoded and combined with the corresponding features. A cross-attention mechanism then models temporal relationships, using the target date as query and the Sentinel-1 and Sentinel-2 features as keys and values. Finally, a probabilistic decoder predicts the optical image and an associated uncertainty map.}
\label{fig_model}
\end{figure*}

The proposed architecture, illustrated in Fig.~\ref{fig_model}, is specifically designed to address this conditional generation task under sparse and irregular multimodal observations. It jointly exploits Sentinel-2 RGB–NIR bands and Sentinel-1 VV/VH radar measurements, without relying on any external masks (e.g., cloud masks).

The model is structured into three main components. First, a 2D spatial feature extraction module independently processes each Sentinel-1 and Sentinel-2 acquisition using convolutional encoders to capture spatial patterns within each image. Second, a temporal encoding and cross-attention module handles the temporal dimension by encoding acquisition dates and selectively aggregating information across time from multimodal observations, enabling the network to operate on variable-length and unordered time series. Finally, a probabilistic decoder generates the optical image and its associated uncertainty at the target date from the aggregated spatio-temporal representation.

\subsubsection{Spatial Feature Extraction}

The first module of the proposed architecture is dedicated to spatial feature extraction from Sentinel-1 and Sentinel-2 observations. Separate convolutional encoders are used for each modality, as optical and radar data convey fundamentally different types of information and spatial patterns. Empirically, this design was found to perform better than a shared encoder, as it allows each branch to learn modality-specific representations while preserving a consistent architectural structure. Although the two encoders share the same architecture, their weights are learned independently.

Spatial feature extraction is performed exclusively in two dimensions. The input time series, of shape $B \times T \times C \times H \times W$ are reshape into $ (B \times T) \times C \times H \times W$ so that each acquisition is processed independently. This design ensures that spatial representations are learned at the image level, without introducing any temporal modeling at this stage.

Each encoder is based on a convolutional backbone equipped with a Spatial Pyramid Pooling (SPP \cite{SPP}) mechanism and outputs spatial features at half the input resolution. The SPP module aggregates multi-scale contextual information through pooling operations at different spatial scales, while maintaining the original feature map resolution via subsequent upsampling and concatenation. This design enables the integration of contextual information across multiple spatial scales while preserving fine-grained spatial details and limiting overly smooth predictions. The resulting feature representation provides a compact and informative spatial description of each acquisition, which is subsequently enriched with temporal information in the next stage of the architecture.

\subsubsection{Temporal Encoding}

The second module of the architecture is dedicated to modeling the temporal dimension of the multimodal satellite time series. Modeling the temporal dimension is critical when working with sparse and irregular satellite observations, as acquisition intervals are highly variable and images acquired closer in time are generally more similar than those separated by longer gaps. In addition, land surface dynamics often exhibit seasonal patterns, meaning that the acquisition date itself provides important contextual information. Explicit temporal encoding therefore allows the model to account for both temporal proximity and seasonal regularities when interpreting multimodal observations, and has been shown to play a key role in sequence-to-sequence image modeling, where removing positional encoding results in a substantial degradation of reconstruction performance \cite{U-TILISE}.

To this end, the acquisition date of each Sentinel-1 and Sentinel-2 image is explicitly encoded and incorporated into the network. Temporal information is modeled using two complementary encodings, distinguishing the target prediction date from the input acquisition dates. For the target date $d_{target}$, we build a continuous temporal representation based on the calendar position within the year, capturing seasonal information:

\begin{equation}
Encode(d) = [y_{d} - y_{0}, \sin(2\pi DOY(d)), \cos(2\pi DOY(d))]
\label{encode_time_relatif}
\end{equation}

where $y_{d}$ denotes the acquisition year, $DOY(d)$ the normalized day of the year, and $y_{0}$ a reference year. This date representation is then projected through a lightweight MLP to obtain the target temporal token used as the query in cross-attention mechanism.

For each input acquisition date $d_{i}$, temporal information is encoded relative to the target date by computing the normalized signed time difference $\Delta{d}= d_{i} - d_{traget}$. This relative offset is encoded using the same sinusoidal temporal encoding as the target date:

\begin{equation}
Encode(d_{i}, d_{target}) = [\Delta{d}, \sin(2\pi \Delta{d}), \cos(2\pi \Delta{d})]
\label{encode_time_absolue}
\end{equation} 

This relative temporal encoding explicitly captures both the temporal distance and direction (past or future) of each observation with respect to the target date. After projection through a lightweight MLP, it is combined with the corresponding spatial features to form spatio-temporal tokens that jointly encode image content and acquisition time.

\subsubsection{Cross-Attention}

To explicitly exploit temporal dependencies in the spatio-temporal representations, we rely on a cross-attention mechanism that conditions the aggregation of input features on the target date. Since the objective is to generate a single optical image at a specified target date, this design focuses on learning a target-conditioned representation rather than modeling the full temporal dynamics, allowing the model to focus on the most temporally relevant observations for generating the optical image.

The spatio-temporal features are first rearranged into a per-pixel token representation. Specifically, features are reshaped into tensors of size $ B \times (H \times W) \times T_{S2 + S1} \times d_{features}$, such that each spatial location is associated with a temporal sequence of feature vectors. To explicitly distinguish between Sentinel-1 and Sentinel-2 observations, a modality encoding is concatenated to each token, allowing the network to identify whether a given feature originates from radar or optical data. The spatio-temporal tokens from both modalities are then concatenated along the temporal dimension and provided as input to the cross-attention module.

To condition the generation process on the target date, we employ a cross-attention mechanism that selectively aggregates information from the available spatio-temporal observations. The temporal encoding of the target date acts as the \emph{query}, while the concatenated multimodal spatio-temporal tokens form the \emph{keys} and \emph{values}. This mechanism is implemented as a stack of cross-attention layers, which allow the model to progressively refines the latent representation by selectively attending to the most temporally relevant observations. No explicit cloud or quality masks are provided to the attention mechanism: instead, the model is trained end-to-end to learn how to weight and selectively ignore unreliable observations based on the generation objective.

The cross-attention module outputs a latent representation conditioned on the target date, which integrates multimodal spatio-temporal information for downstream generation. This representation is subsequently passed to the probabilistic decoder to generate the predicted optical image and its associated uncertainty.

\subsubsection{Probabilistic Decoder}

The target-date–conditioned latent representation is fed to a probabilistic decoder that predicts the parameters of a pixel-wise conditional distribution, enabling explicit estimation of predictive uncertainty together with the generated optical image.

Specifically, the decoder estimates the parameters of a Laplace distribution independently for each spectral band and spatial location. A Laplace distribution is preferred over a Gaussian one, as its heavier-tailed formulation is more robust to large prediction errors and was empirically observed to reduce over-smoothing effects in the generated images. For a given pixel and channel, the predicted distribution is defined by a mean parameter $\mu$ and a scale parameter $b>0$, where the latter controls the dispersion of the distribution. The decoder outputs $\mu$ and the logarithm of the scale parameter $\log(b)$ to ensure numerical stability during training. Importantly, although the formulation is probabilistic, the inference process itself is fully deterministic: for a given input, the model always predicts the same distribution parameters, without any stochastic sampling.

Under this Laplace assumption, learning is formulated through the minimization of the corresponding negative log-likelihood. For a pixel $i$ and a spectral channel $c$, the loss term is expressed as:

\begin{equation}
\mathcal{L}_{Laplace} = \frac{| y_{i, c} - \mu_{i,c} |}{b_{i,c}} + log(2b_{i,c})
\label{nll_laplace}
\end{equation}

This objective naturally balances reconstruction accuracy and uncertainty estimation: pixels associated with higher prediction errors can be assigned larger uncertainty, while confident predictions are encouraged to remain sharp.

\section{Data}

A dedicated dataset was assembled for this study using Sentinel satellite observations retrieved through the GEODES platform. The dataset integrates Sentinel-2 Level-2A optical data and Sentinel-1 Ground Range Detected (GRD) radar data, whose complementary characteristics are well suited for land surface monitoring.

For Sentinel-2, only the RGB and near-infrared (NIR) bands were used. The data were converted to surface reflectance by dividing the digital values by $10.000$, and subsequently clipped to the range $[0, 1]$. No additional normalization or correction was applied in order to preserve the original radiometric information as much as possible.

Sentinel-1 data were processed using Ground Range Detected (GRD) products, considering only the VV and VH polarizations. Only a single acquisition orbit was retained, the descending orbit, in order to ensure consistent viewing geometry and minimize radiometric variability across the time series. The radar images were orthorectified and aligned to the Sentinel-2 tiling system, and speckle filtering was performed using the S1Tiling tool box \cite{S1TILING}. The values were then converted to decibel scale and standardized using the mean and standard deviation computed over the dataset.

\begin{figure}[htbp]
\centering
\includegraphics[width=0.25\textwidth]{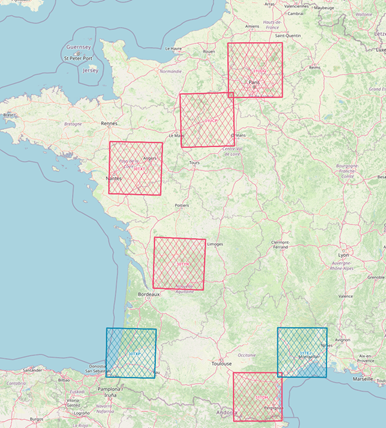}
\caption{Geographical distribution of the training (pink) and evaluation (blue) tiles used in this study.}
\label{fig_dataset}
\end{figure}

Fig.~\ref{fig_dataset} illustrates the geographical distribution of the tiles used to build the training dataset and the independent evaluation dataset. The selected tiles are well distributed across both northern and southern regions, ensuring a diverse range of landscapes and acquisition conditions. 

Based on these selected tiles, temporal series were downloaded with particular care to cover all seasons. A total of $25.813$ patches of size $96 \times 96$ pixels were extracted for training. Each patch consists of a temporal sequence of 8 images, including temporally close Sentinel-1 and Sentinel-2 acquisitions. The dataset is largely composed of agricultural areas, which were intentionally favored due to their strong temporal variability. Patches containing water bodies were excluded from the training dataset.

\section{Experiments}

\subsection{Training and implementation}

A self-supervised masking strategy is adopted to train the model. For each Sentinel-1 / Sentinel-2 temporal sequence patch extracted from the dataset, one Sentinel-2 patch is deliberately removed and used as the prediction target. Unlike cloud-removal approaches that focus on partial pixel-wise reconstruction, the objective here is to generate a complete optical image. The masked optical image is randomly selected either between two available frames (interpolation setting) or as the last image of the sequence (extrapolation setting), allowing the model to learn both reconstruction within the temporal range and forecasting beyond the last observation.

To further improve robustness to sparse and irregular acquisitions, the length of the input time series is randomly reduced during training. This encourages the network to operate under varying temporal contexts and prevents over-reliance on fixed sequence lengths. Importantly, no temporal padding is applied: the model is trained directly on sequences of variable length.

The model is trained by minimizing the negative log-likelihood associated with the Laplace output distribution, as defined in \eqref{nll_laplace}.

\subsection{Evaluation metrics}

The performance of the proposed method is assessed using standard pixel-wise error metrics, namely the Mean Absolute Error (MAE), the Root Mean Squared Error (RMSE), and the Peak Signal-to-Noise Ratio (PSNR). All metrics are computed exclusively on non-cloudy pixels, as identified by the cloud masks provides by MAJA \cite{MAJA}. Metrics are evaluated on the four Sentinel-2 spectral bands corresponding to the RGB and near-infrared (NIR) channels. Inference is performed on large-scale scenes (typically $2000 \times 2000$ or $4000 \times 4000$ pixels, depending on the experiment), and metrics are then computed on the full images. This evaluation protocol not only measures pixel-level accuracy, but also assesses the model’s ability to generate spatially consistent predictions at large scale. In addition, land-cover masks derived from CES OSO products \cite{OSO} are used to analyze error distributions across different surface types and to identify areas where the model is more prone to inaccuracies. In interpolation settings, the proposed method is compared with both a simple linear interpolation baseline, commonly used in Sentinel-2 time series reconstruction studies, and U-TILISE \cite{U-TILISE}, a sequence-to-sequence model based on a convolutional encoder–decoder architecture that operates exclusively on optical imagery. For a fair comparison, U-TILISE was retrained on our dataset using the authors’ publicly available code and recommended training protocol.

Beyond point-wise accuracy, the quality of the predicted uncertainty is evaluated using calibration curves, which measure the empirical coverage of prediction intervals against their nominal confidence levels.

\subsection{Setup}

All experiments are conducted using real Sentinel-1 and Sentinel-2 satellite images. For each experiment, a target date is randomly selected among dates for which a cloud-free Sentinel-2 image is available and used as ground truth. A significant degree of randomness is introduced in the selection of both target dates and input acquisitions, ensuring that the model is evaluated across a wide range of temporal configurations rather than fixed or short time intervals.

Depending on the experimental setting, the construction of the multimodal input sequence follows different temporal constraints. In the interpolation setting, only acquisitions that temporally surround the target date are selected as inputs. In contrast, in the extrapolation setting, all input acquisitions are strictly located before the target date. In both cases, the target Sentinel-2 observation is never included among the optical inputs provided to the model.

The resulting input sequences consist of Sentinel-2 (cloudy or non-cloudy) and Sentinel-1 images selected according to these temporal constraints. No filtering is applied based on the cloud coverage percentage of the Sentinel-2 inputs, meaning that heavily cloud-contaminated observations may be provided to the model. Sentinel-1 and Sentinel-2 data are not temporally aligned, and depending on data availability, a Sentinel-1 acquisition may coincide with the target date to be generated.
:
When multiple target dates are considered from the same input observations, the model is applied independently to each date. The multimodal input sequence remains unchanged, while only the target date encoding is updated, resulting in $N$ forward passes to generate $N$ images.

To avoid trivial reconstruction scenarios, a minimum temporal gap is randomly enforced between the target date and the closest Sentinel-2 input acquisitions. This forced temporal offset prevents the network from simply copying the nearest Sentinel-2 observation and ensures that it must genuinely infer temporal dynamics. The maximum sequence length is fixed to $8$ images for all experiments, however the actual sequence length may vary depending on data availability and the selected target date.

\section{Results}

\subsection{Interpolation}

The quantitative results are reported in Tab.~\ref{tab_comparison_ai_linear}. Overall, the proposed model achieves the best performance across all evaluated metrics and land-cover classes. Compared to classical linear interpolation, the gains remain moderate for relatively stable classes such as urban areas and forests. This behavior is expected, as these land-cover types typically exhibit limited temporal variability between successive observations, making them reasonably well suited to simple interpolation strategies. Urban areas nevertheless present slightly higher error levels than forests, as Sentinel-2 reflectance values can be highly saturated in urban environments.

In contrast, larger performance gaps are observed for cropland areas, which are characterized by strong seasonal dynamics and pronounced temporal variations. While linear interpolation struggles to model these non-linear temporal patterns (MAE of 0.018 and RMSE of 0.030), the proposed model significantly reduces the errors (MAE of 0.016 and RMSE of 0.025), highlighting its ability to capture complex temporal evolution.

A similar trend is observed when comparing our approach to U-TILISE model. Across all land-cover classes, our model achieves lower MAE and RMSE values and higher PSNR. The difference is again particularly marked for cropland, where U-TILISE reaches an MAE of 0.022 and an RMSE of 0.035. It is important to emphasize that U-TILISE relies exclusively on optical Sentinel-2 observations, whereas our model leverages both Sentinel-1 and Sentinel-2 data. The complementary structural information provided by SAR, which is insensitive to cloud cover and illumination conditions, likely contributes to the systematic improvements observed here.

Taken together, these results indicate that the proposed multimodal approach preserves robustness in stable environments while providing clear benefits in highly dynamic settings, where accurate temporal reconstruction is inherently more challenging.

\begin{table*}[t]
\centering
\caption{Quantitative evaluation of different time series reconstruction methods in interpolation mode.}
\label{tab_comparison_ai_linear}
\begin{tabular}{lccccccccc}
\hline
\multirow{2}{*}{Method} 
& \multicolumn{3}{c}{MAE $\downarrow$} 
& \multicolumn{3}{c}{RMSE $\downarrow$} 
& \multicolumn{3}{c}{PSNR $\uparrow$} \\
\cline{2-10}
& Urban & Forest & Cropland
& Urban & Forest & Cropland
& Urban & Forest & Cropland \\
\hline
Ours
& \textbf{0.012} & \textbf{0.007} & \textbf{0.016}
& \textbf{0.018} & \textbf{0.011} & \textbf{0.025}
& \textbf{34.841} & \textbf{38.726} & \textbf{32.201} \\
Linear interpolation 
& 0.012 & 0.008 & 0.018
& 0.018 & 0.015 & 0.030
& 34.675 & 37.432 & 30.790 \\
U-TILISE
& 0.015 & 0.011 & 0.022
& 0.022 & 0.018 & 0.035
& 33.242 & 35.535 & 29.904 \\

\hline
\end{tabular}
\end{table*}

\begin{figure*}[htbp]
\centering
\includegraphics[width=0.8\textwidth]{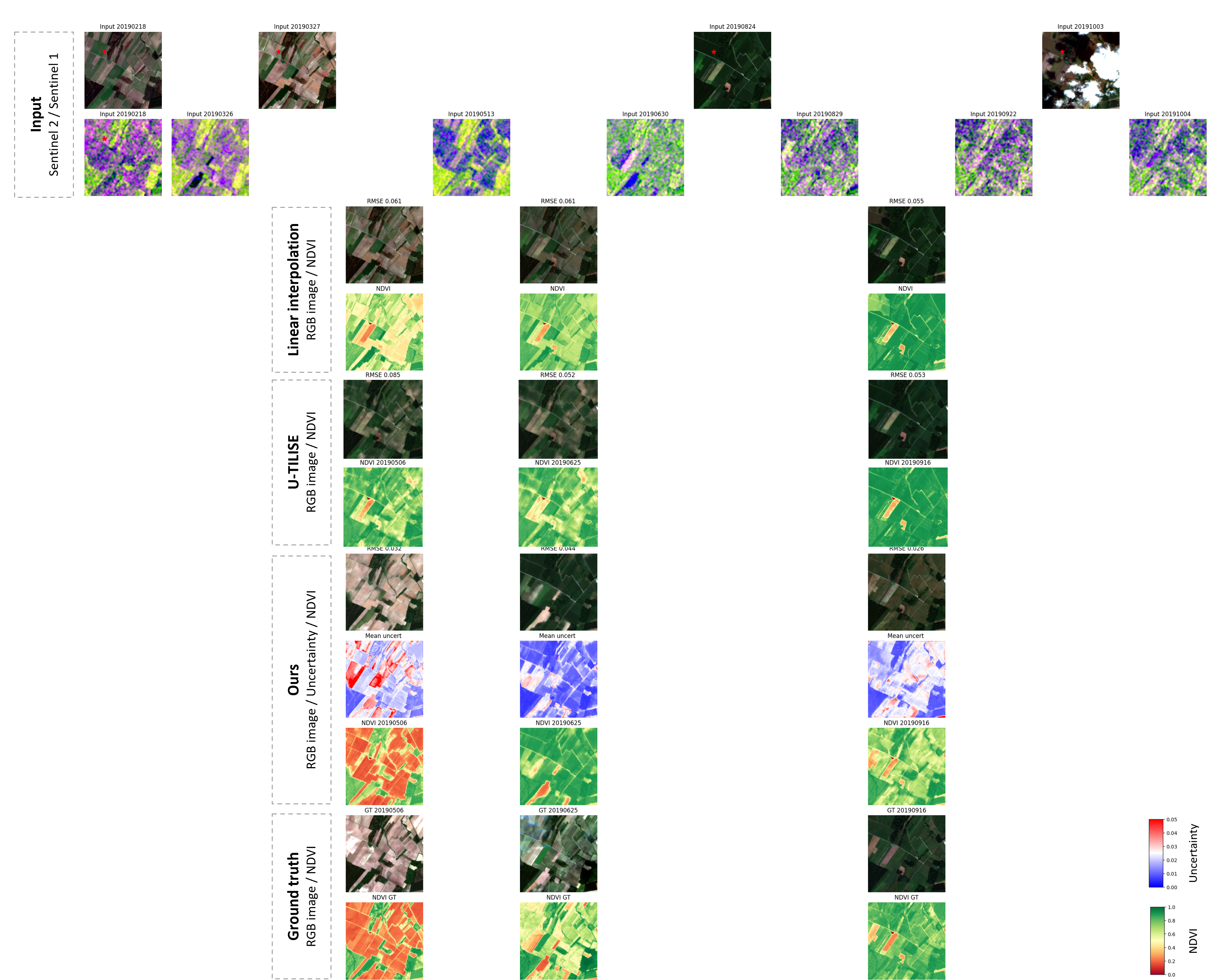}
\caption{Qualitative comparison of large-gap Sentinel-2 image reconstruction using linear interpolation, U-TILISE and the proposed method. Input Sentinel-2 and Sentinel-1 images are shown at the top (chronologically ordered), followed by the reconstructed RGB images and NDVI obtained with linear interpolation, U-TILISE, the proposed method (including uncertainty maps), and the corresponding Sentinel-2 ground truth.}
\label{fig_nn_vs_linear}
\end{figure*}

Fig.~\ref{fig_nn_vs_linear} presents a qualitative comparison between the proposed model and the different time series reconstruction methods in a challenging scenario characterized by a very limited number of available observations and large temporal gaps between the input dates and the generated target dates, with an average offset of approximately one month and up to two months in the most extreme cases. The scene corresponds to an area with strong vegetation dynamics. In this context, the proposed model effectively leverages Sentinel-1 information, which provides complementary cues related to phenological changes, enabling a more accurate reconstruction of vegetation evolution. In contrast, linear interpolation and U-TILISE, both relying solely on optical information, are less able to capture these dynamics, resulting in a less accurate reconstruction of the underlying phenological patterns. When the temporal gap becomes extremely large, as observed for the first generated image in the sequence, the prediction produced by the proposed model appear smoother and less detailed. Importantly, these regions also exhibit higher predicted uncertainty, indicating that the model appropriately reflects its reduced confidence in areas where the temporal extrapolation is most challenging. This behavior is desirable, as it suggests that the uncertainty estimates are coherent with the actual difficulty of the reconstruction task.

\begin{figure}[htbp]
\centering
\includegraphics[width=0.5\textwidth]{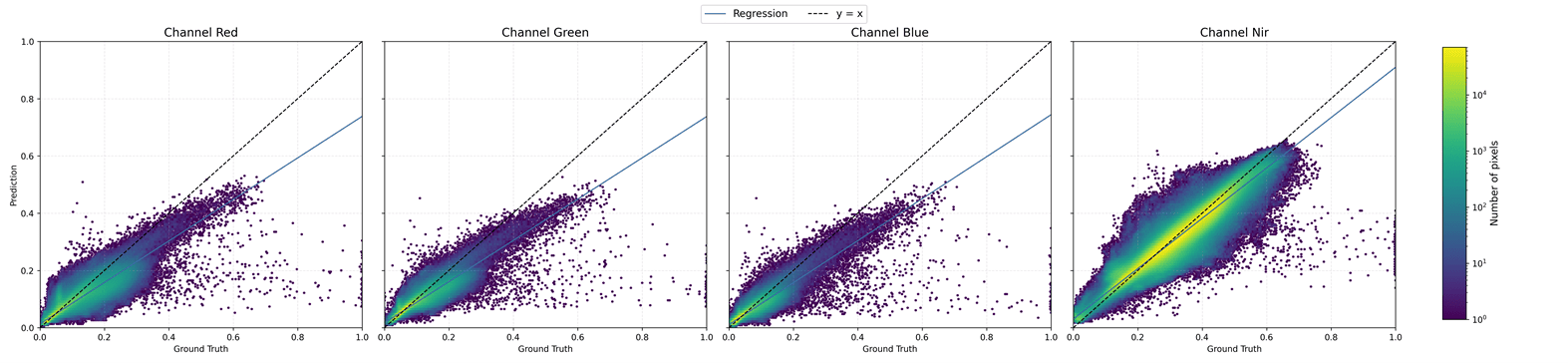}
\caption{Channel-wise comparison of ground-truth and predicted reflectances obtained in interpolation mode. From left to right: B4 (red), B3 (green), B2 (blue), and B8 (near-infrared).}
\label{fig_reflectance_eval_intern}
\end{figure}

Fig.~\ref{fig_reflectance_eval_intern} shows scatter density plots comparing reflectance values predicted by the proposed method with the corresponding ground-truth Sentinel-2 reflectances. Each point represents a non-cloudy pixel, and colors indicate pixel density. The dashed black line corresponds to the ideal line, while the solid line shows the linear regression fitted to the predictions. Overall, a strong correlation is observed across all channels, indicating that the model preserves the radiometric structure of the observations. Slight deviations from the identity line are visible, particularly at higher reflectance values, where predictions tend to be smoother. This effect is more pronounced in the NIR band, which is known to present larger variability and stronger sensitivity to vegetation dynamics. Nevertheless, the concentration of points along the identity line confirms the model’s ability to produce radiometrically consistent predictions over a wide range of reflectance values.

\begin{figure}[htbp]
\centering
\includegraphics[width=0.3\textwidth]{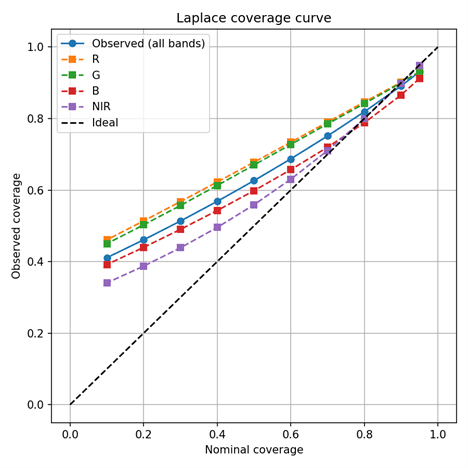}
\caption{Calibration curves (Laplace coverage) for the predicted reflectance uncertainty across spectral bands.}
\label{incertitude_interpolation}
\end{figure}

Fig.~\ref{incertitude_interpolation} presents the calibration curves derived from the Laplace predictive distributions, comparing nominal and empirical coverage across spectral bands. The curves remain close to the ideal diagonal over the full range of coverage levels, indicating that the predicted uncertainty intervals are well aligned with the true error distribution. Slight over-coverage is observed at lower nominal levels, particularly for the green band, where the empirical coverage slightly exceeds the nominal one. This indicates that the predicted intervals are slightly wider than required at these coverage levels. As the nominal coverage increases, all spectral bands converge toward the ideal line, demonstrating stable calibration behavior at higher confidence levels. The aggregated “all bands” curve closely follows the diagonal, confirming consistent uncertainty quantification across spectral channels.

\subsection{Extrapolation}

Tab.~\ref{extrapolation_errors} reports quantitative results in extrapolation mode on valid pixels (non-water, non-cloudy) and cropland areas. We restrict the analysis to these two categories, as cropland represents the most relevant and challenging class, while valid pixels provide a global assessment. As expected in an extrapolation setting, the proposed method exhibits higher reconstruction errors compared to simple  interpolation. Nevertheless, the absolute error values remain within reasonable and physically meaningful ranges, indicating that the model maintains a good level of radiometric consistency despite the increased difficulty of predicting beyond the temporal support of the observations.

\begin{table}[t]
\centering
\caption{Quantitative performance of the proposed method in extrapolation mode on cropland and valid pixels (non-water, non-cloudy).}
\label{extrapolation_errors}
\begin{tabular}{cc|cc|cc}
\hline
\multicolumn{2}{c|}{MAE $\downarrow$} 
& \multicolumn{2}{c|}{RMSE $\downarrow$} 
& \multicolumn{2}{c}{PSNR $\uparrow$} \\
\cline{1-6}
Cropland & Valid
& Cropland & Valid
& Cropland & Valid \\
\hline
0.027 & 0.022
& 0.040 & 0.034
& 28.323 & 29.735 \\
\hline
\end{tabular}
\end{table}

\begin{figure}[htbp]
\centering
\includegraphics[width=0.5\textwidth]{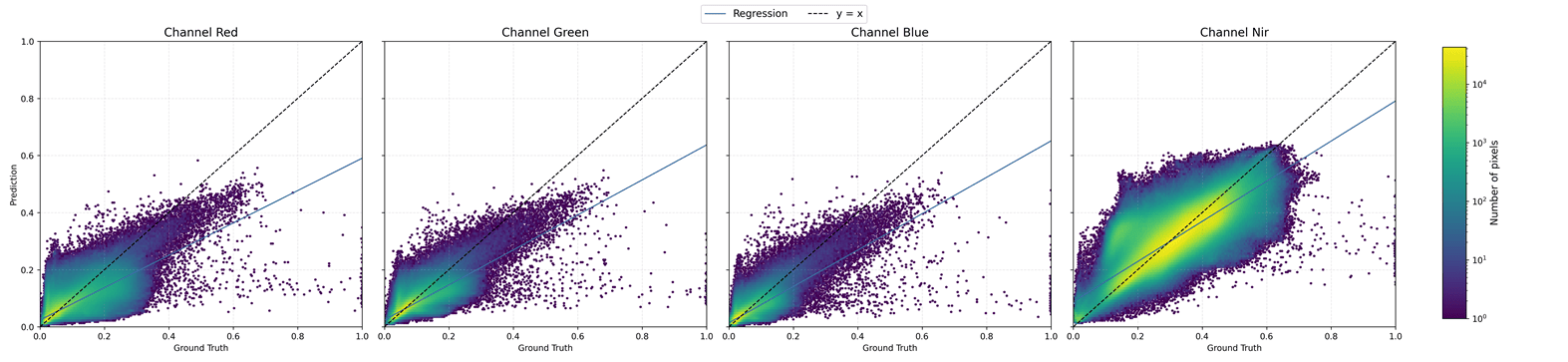}
\caption{Channel-wise comparison of ground-truth and predicted reflectances obtained in extrapolation mode. From left to right: Red, Green, Blue, Near-infrared.}
\label{fig_reflectance_eval_extrap}
\end{figure}

Compared to the interpolation setting, the extrapolation results show a larger dispersion around the identity line across all spectral bands (Fig.~\ref{fig_reflectance_eval_extrap}), indicating reduced radiometric accuracy. This behavior is expected, as extrapolation is inherently more challenging than interpolation, requiring predictions outside the temporal support of cloud-free observations. Despite this increased difficulty, a clear correlation between predicted and ground-truth reflectances is preserved, although deviations from the ideal line are more pronounced at higher reflectance values, particularly in the NIR band. Overall, the results highlight the performance gap between interpolation and extrapolation while confirming the preservation of consistent radiometric trends.

\begin{figure*}[htbp]
\centering
\includegraphics[width=0.8\textwidth]{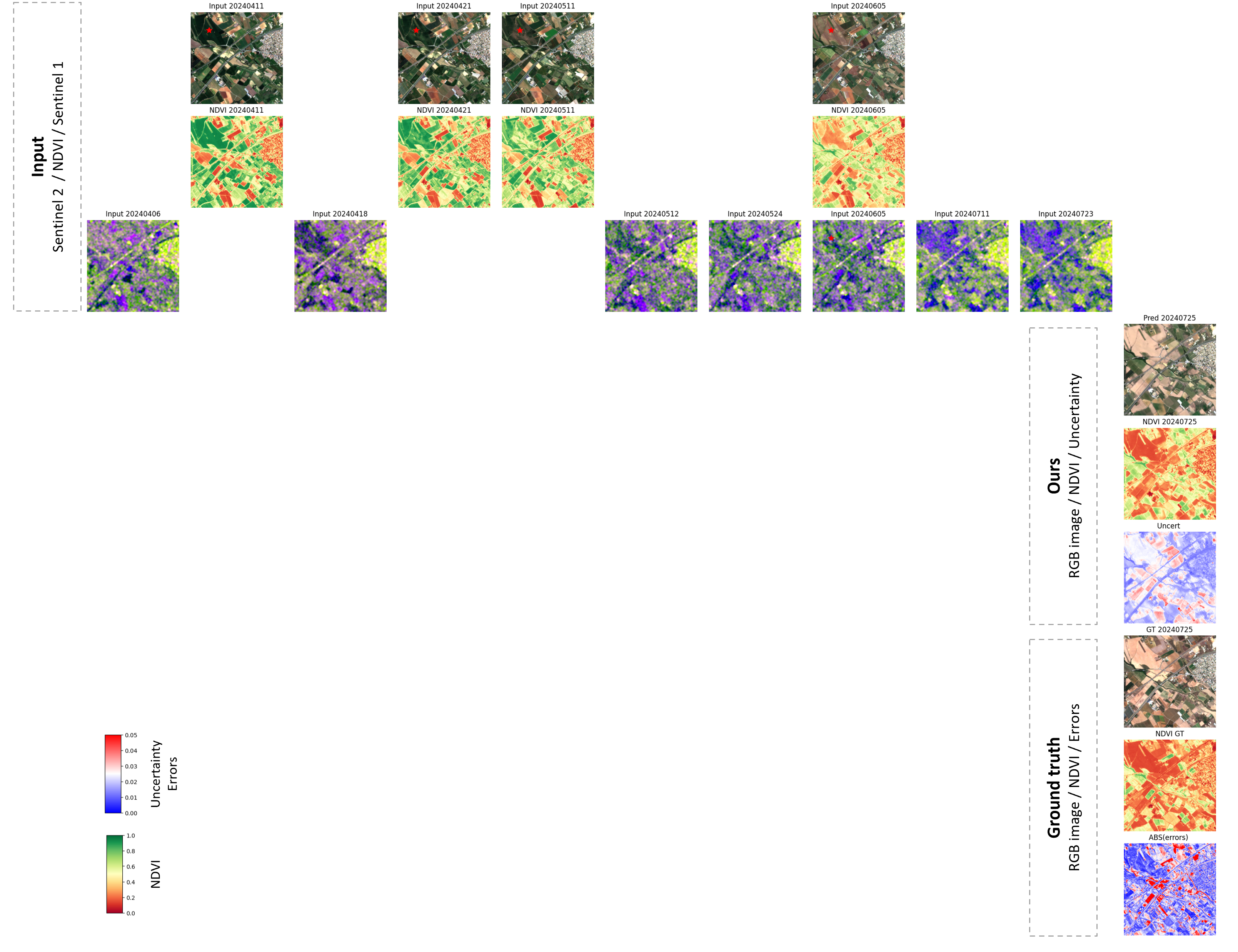}
\caption{Qualitative evaluation of long-term Sentinel-2 image extrapolation. Input Sentinel-2 (RGB / NDVI) and Sentinel-1 images are shown at the top (chronologically ordered), followed by the extrapolated RGB image, NDVI, associated uncertainty map, the corresponding Sentinel-2 ground truth and the absolute difference map.
}
\label{fig_extra_visu}
\end{figure*}

Fig.~\ref{fig_extra_visu} illustrates a challenging extrapolation scenario in which the model is tasked with generating a Sentinel-2 optical image 1 month and 20 days after the last available cloud-free optical observation. Despite the long temporal gap, the model is able to capture the main vegetation changes occurring at the field scale, as reflected by the evolution of NDVI between the input observations, the prediction, and the ground truth. Furthermore, the predicted uncertainty shows a strong visual correlation with the spatial distribution of errors, with higher uncertainty values observed in areas where higher errors are observed.

\section{Experiments}

\subsection{Dense Annual Time Series Reconstruction}

The objective is to reconstruct a dense and fully cloud-free optical time series over a one-year period. The model is evaluated on a complete annual sequence by generating images at regular temporal intervals of 5 days for dense NDVI analysis, and 20 days for visual inspection of the reconstructed images. In this setting, the network is required to generate a one-year synthetic optical time series by producing images at target dates for which no optical observations are available, as well as re-generating images at dates where optical data exist, in order to remove possible cloud contamination. This setting allows us to evaluate how well the model handles cloud-contaminated observations. At each inference step, the model operates on a sliding temporal window limited to a maximum of $8$ Sentinel-2 acquisitions. Since the model generates a single image per inference, the dense annual time series is obtained by independently applying the model to each target date.

\begin{figure*}[htbp]
\centering
\includegraphics[width=1\textwidth]{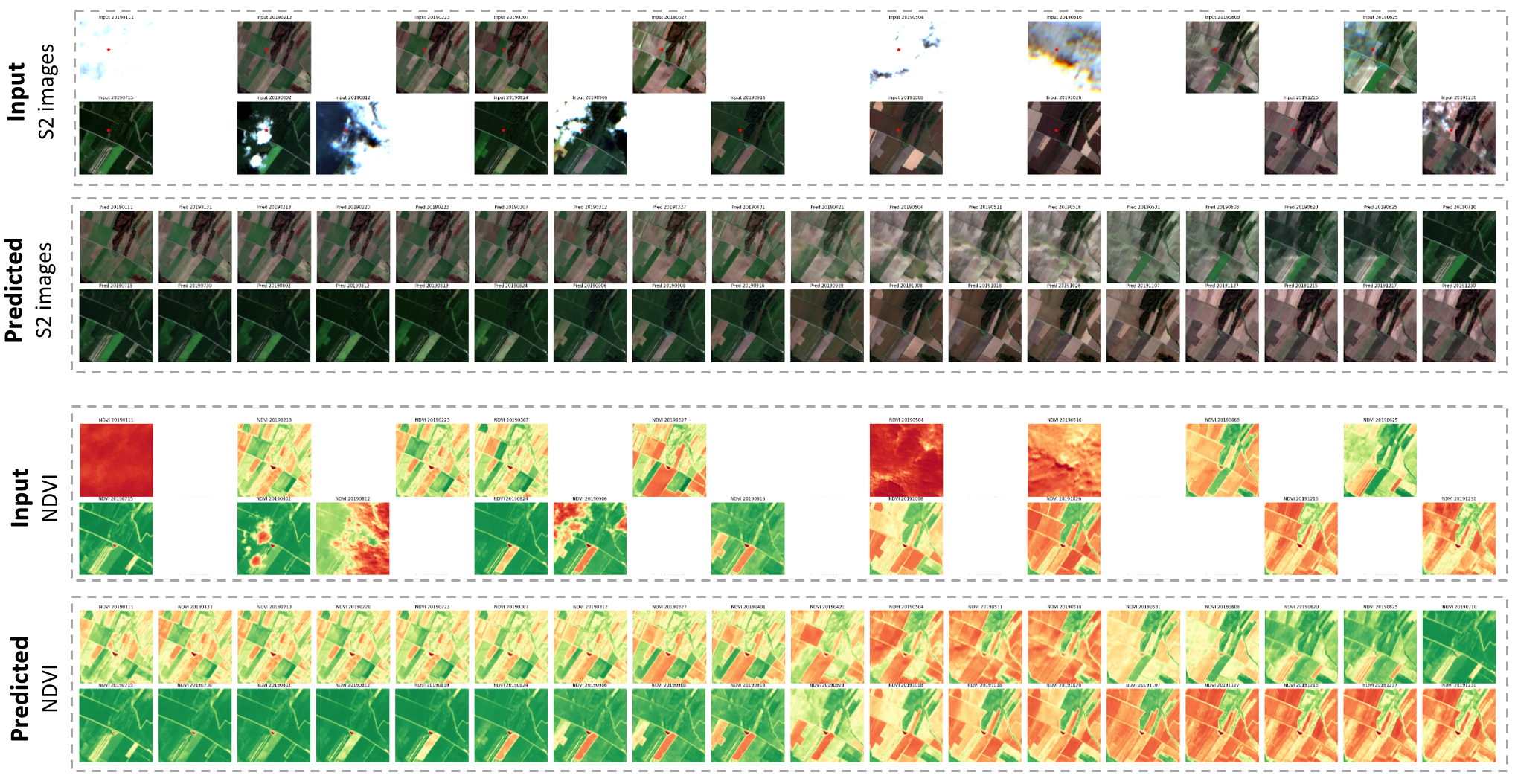}
\caption{Reconstruction of a one-year Sentinel-2 time series and associated NDVI from sparse and cloudy observations.}
\label{dense_img}
\end{figure*}

Fig.~\ref{dense_img} illustrates the behavior of the model under this evaluation setting. The top row displays the exact Sentinel-2 optical acquisitions used as input, which constitute the only optical observations available for time series reconstruction. Several consecutive dates are severely affected by dense cloud cover, resulting in highly irregular and noisy optical time series. Despite these challenging conditions, the reconstructed images remain sharp and well defined, preserving fine spatial structures and avoiding the overly smooth or blurry artifacts commonly associated with temporal interpolation.

Notably, the model is not provided with any explicit cloud or quality masks and is never trained to detect or remove clouds. Nevertheless, the predictions indicate that the network is able to implicitly identify and suppress cloud-covered regions, effectively learning an internal masking mechanism to restore missing or corrupted information. This behavior is clearly reflected in the generated images, where the temporal evolution of vegetation is visually consistent across dates and exhibits a smooth and well-preserved NDVI progression, in contrast to the abrupt and unstable variations observed in the input acquisitions.

\begin{figure}[htbp]
\centering
\includegraphics[width=0.4\textwidth]{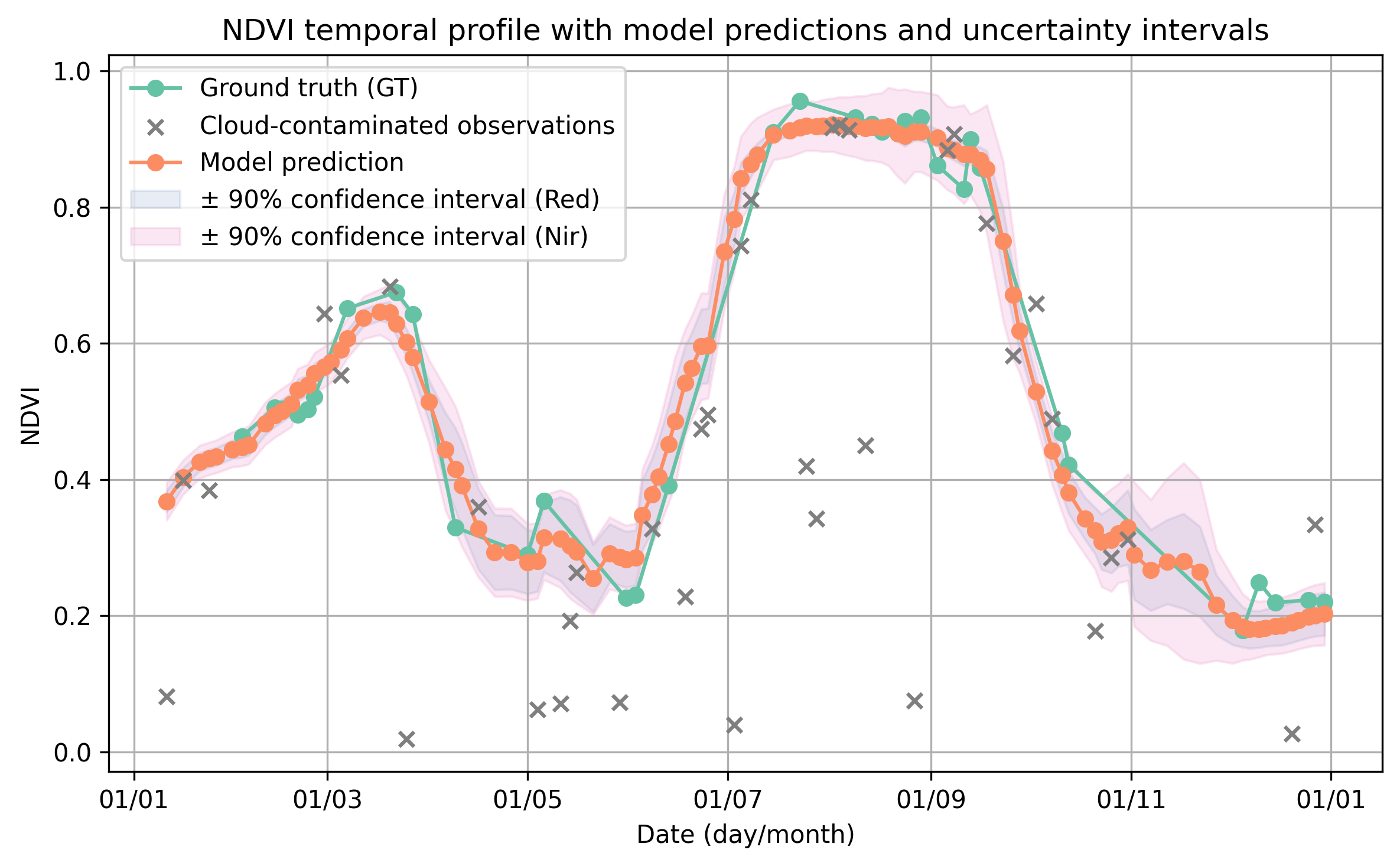}
\caption{One-year pixel-wise NDVI time series reconstruction with 5-day predictions and uncertainty.
NDVI time series for the pixel marked by a red star in Fig.~\ref{dense_img}, including model predictions and uncertainty intervals. Predictions are generated every 5 days over a full one-year period.}
\label{dense_NDVI_curve}
\end{figure}

Fig.~\ref{dense_NDVI_curve} presents the NDVI temporal profile extracted from the pixel marked by a red star in Fig.~\ref{dense_img} (top row), comparing ground-truth observations with model predictions and associated uncertainty intervals over a full one-year period. Predictions are generated every 5 days, resulting in a dense reconstruction of the NDVI despite sparse and cloudy optical observations. The predicted NDVI closely follows the ground-truth dynamics throughout the year, accurately capturing the main phenological phases of vegetation, including vegetation growth, peak, and decline.

Importantly, the predicted uncertainty intervals remain well calibrated over time, consistently enclosing the ground-truth NDVI values, particularly during periods of strong cloud contamination and rapid temporal transitions. This indicates that the model not only reconstructs a temporally coherent and physically plausible NDVI signal, but also provides uncertainty estimates that adapt to the observation conditions. Notably, a marked increase in uncertainty is observed around early November (01/11), which corresponds to a period with no available cloud-free Sentinel-2 observations. Consequently, the network relies on limited or indirect information and accordingly expresses higher predictive uncertainty. This behavior reflects a desirable property of the model, as the estimated uncertainty increases when the input data are sparse or less informative, rather than remaining artificially overconfident.

\subsection{Influence of SAR data}

To assess the actual contribution of radar information, we trained an additional model following exactly the same training protocol, architecture, and optimization settings as previously described, but using only Sentinel-2 optical observations as input. The performance of this S2-only model was then compared with that of the multimodal model combining Sentinel-1 (SAR) and Sentinel-2 (optical) data.

The quantitative results reported in Tab.~\ref{tab:sar_comparison} indicate that the multimodal model consistently outperforms the optical-only variant across all evaluated metrics, both when considering all pixels and when focusing specifically on cropland areas. Although the performance gaps remain moderate, the systematic reduction in MAE and RMSE, together with the increase in PSNR, suggests that the integration of Sentinel-1 data provides complementary information that slightly improves reconstruction accuracy.

\begin{table}[h]
\centering
\caption{Comparison of performance (interpolation and extrapolation) between the multimodal (SAR + optical) and optical-only models.}
\begin{tabular}{lccc}
\hline
Model & MAE & RMSE & PSNR \\
\hline
\multicolumn{4}{c}{\textit{All pixels}} \\
\hline
Multimodal & \textbf{\fpeval{round((0.019 + 0.015) / 2, 3)}} 
& \textbf{\fpeval{round((0.031 + 0.025) / 2, 3)} }
& \textbf{\fpeval{round((30.536 + 32.557 ) / 2, 3)}} \\
Optical only & \fpeval{round((0.022 + 0.015) / 2, 3)} 
& \fpeval{round((0.034 + 0.0266) / 2, 3)} 
& \fpeval{round((29.778 + 32.298) / 2, 3)} \\
\hline
\multicolumn{4}{c}{\textit{Crops only}} \\
\hline
Multimodal &\textbf{ \fpeval{round((0.023 + 0.018 ) / 2, 3)}} 
& \textbf{\fpeval{round((0.036 + 0.0303) / 2, 3)} }
& \textbf{\fpeval{round((29.088 + 31.078 ) / 2, 3)}} \\
Optical only & \fpeval{round((0.027 + 0.019) / 2, 3)} 
& \fpeval{round((0.040 + 0.031) / 2, 3)} 
& \fpeval{round((28.271 +  30.862) / 2, 3)} \\
\hline
\end{tabular}
\label{tab:sar_comparison}
\end{table}

\subsection{Learned attention masks}

To better understand how the model leverages multimodal and multitemporal information, we examine the cross-attention maps generated during inference. Attention weights are visualized across several layers and heads, allowing us to analyze how different representational levels contribute to the selection and weighting of available observations. This analysis sheds light on how the network integrates optical and SAR inputs over time to reconstruct the target-date image.

\begin{figure}[htbp]
\centering
\includegraphics[width=0.4\textwidth]{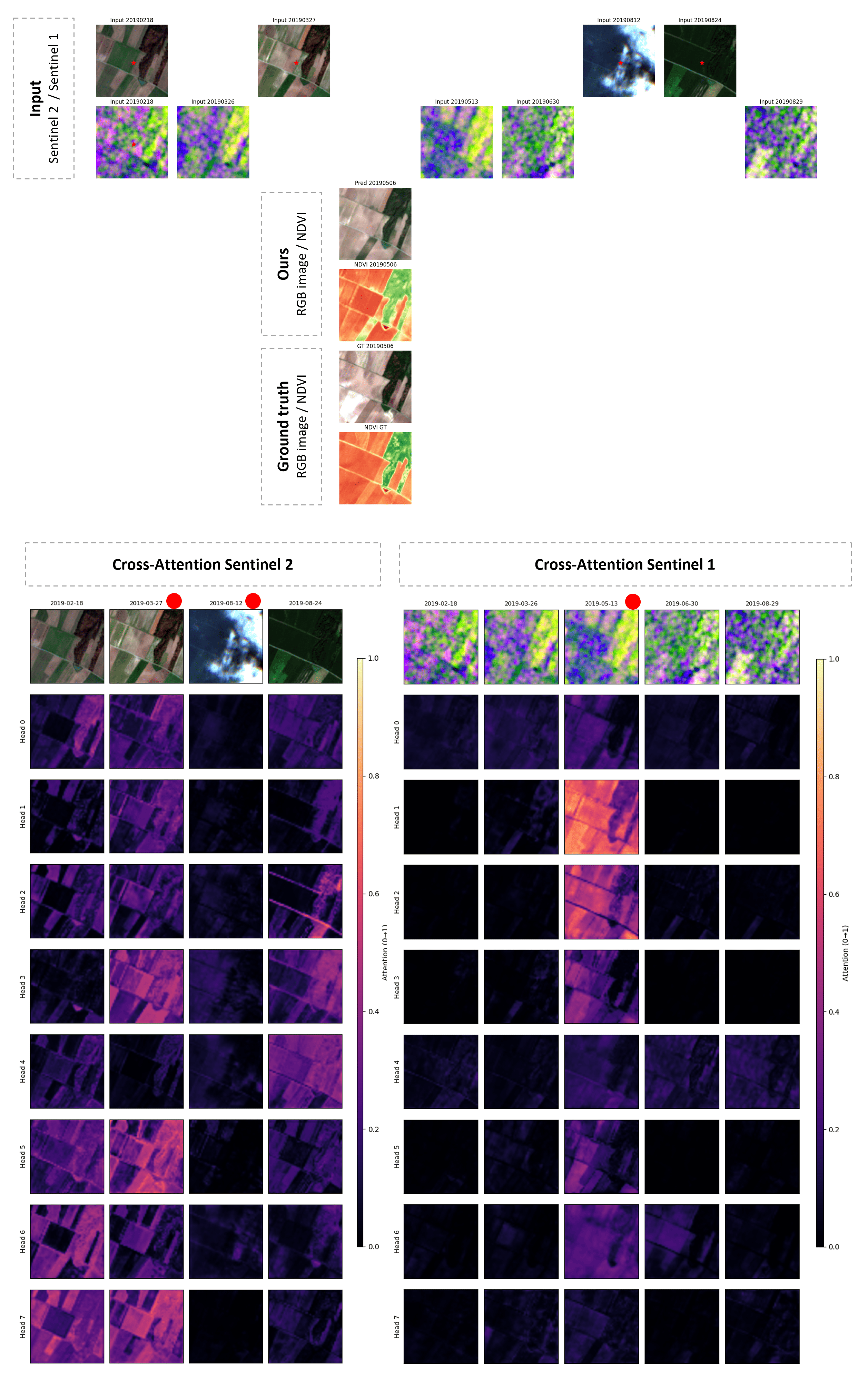}
\caption{Cross-attention analysis under large temporal gap conditions. Top: Sentinel-2 and Sentinel-1 input images, model prediction (RGB and NDVI), and corresponding ground truth (RGB and NDVI).
Bottom: Cross-attention weights associated with each Sentinel-2 and Sentinel-1 input. Red circles indicate the optical acquisitions temporally surrounding the target date and the radar acquisition closest to the target date.}
\label{cross_attention_v1}
\end{figure}

Fig.~\ref{cross_attention_v1} illustrates a challenging reconstruction scenario characterized by a large temporal gap between the target date and the available Sentinel-2 observations provided as input to the model. In this setting, the network must compensate for the absence of temporally close optical observations by exploiting available optical inputs together with Sentinel-1 data to infer the target reflectance. The cross-attention maps confirm that this is indeed the case: the Sentinel-1 acquisition closest to the target date (highlighted with a red circle) receives significantly higher attention weights. This behavior indicates that, in the presence of a large optical temporal gap, the model effectively leverages the most temporally relevant SAR observation to guide the reconstruction. 
Regarding the Sentinel-2 inputs, the optical images closest in time to the target date are also highlighted with red circles. Among them, one acquisition is heavily contaminated by clouds. The cross-attention maps clearly show that the model assigns almost no attention to this cloudy image, despite its temporal proximity to the target date. Interestingly, the cloud pattern can still be faintly perceived in the corresponding attention map.
The model distributes its attention across the remaining cloud-free S2 images acquired before and after the target date. The attention activations appear more spatially diffuse across these acquisitions, suggesting that the network aggregates complementary temporal information rather than relying on a single optical frame. 

\begin{figure}[htbp]
\centering
\includegraphics[width=0.4\textwidth]{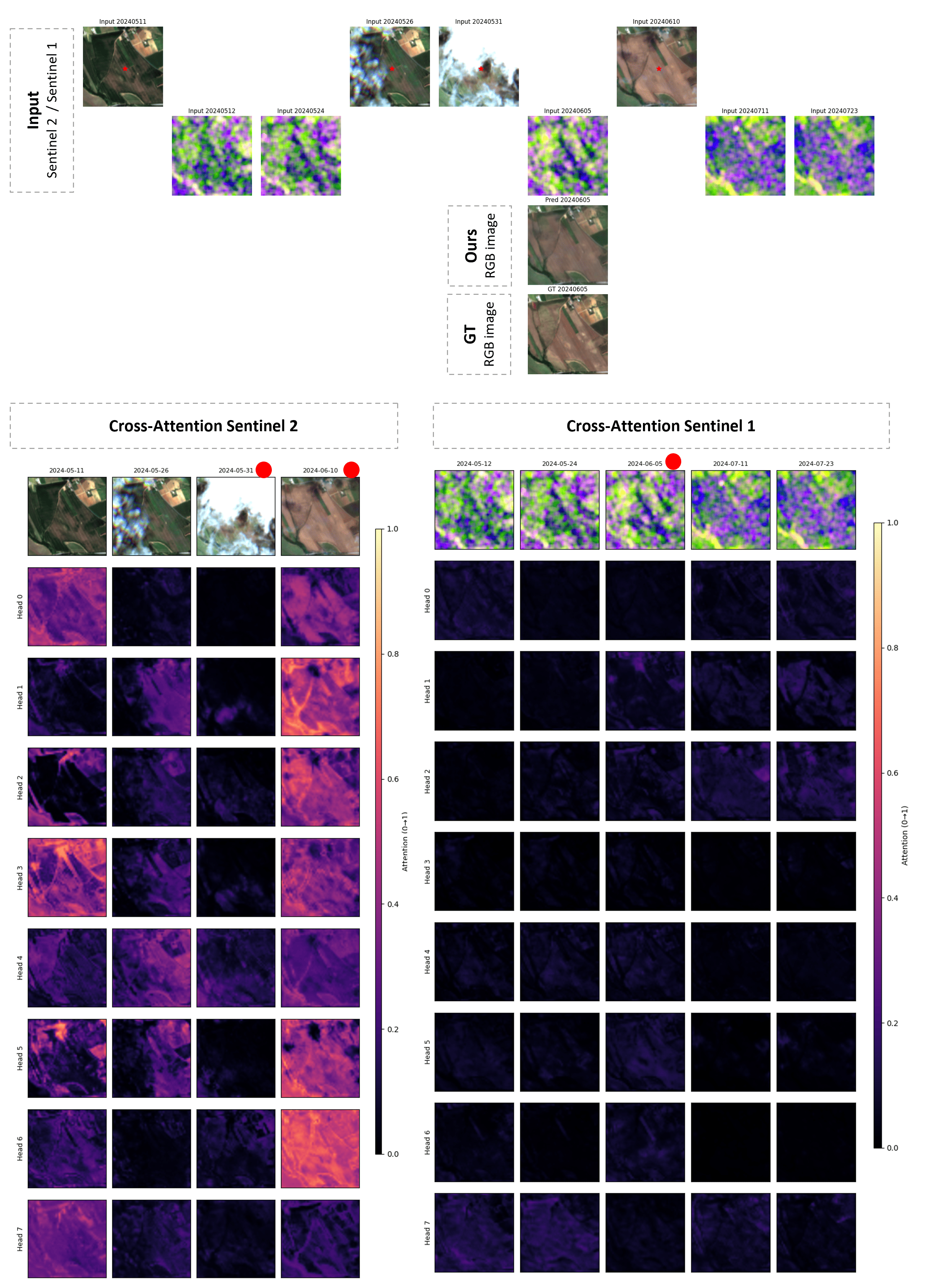}
\caption{Cross-attention analysis in an interpolation setting with temporally proximate optical observations. Top: Sentinel-2 and Sentinel-1 input images, model prediction and corresponding ground truth.
Bottom: Cross-attention weights associated with each Sentinel-2 and Sentinel-1 input. Red circles indicate the optical acquisitions temporally surrounding the target date and the radar acquisition closest to the target date. }
\label{cross_attention_v2}
\end{figure}

Fig.~\ref{cross_attention_v2} presents an interpolation scenario in which the target Sentinel-2 image is generated with a very small temporal gap relative to one of the optical inputs. In this example, the closest S2 acquisition is only five days away from the target date, while the other preceding optical images are heavily contaminated by clouds. In addition, a Sentinel-1 acquisition is available exactly at the target date.
The cross-attention maps show a markedly different behavior compared to the large temporal gap case. Despite the availability of radar data at the target date, the attention weights assigned to the Sentinel-1 inputs are nearly zero across all heads. This indicates that the model does not rely on SAR information in this configuration. Instead, it appears to detect that temporally close and reliable optical information is available and therefore concentrates almost exclusively on the corresponding S2 acquisition. In other words, the model implicitly considers that the relevant information required for reconstruction is already contained in the nearby optical observation and does not require additional structural cues from radar data.
This behavior is further confirmed by the cross-attention maps over the Sentinel-2 inputs. The optical image closest to the target date (marked with a red circle) clearly dominates the attention distribution, receiving higher weights than the other acquisitions. As in the previous experiment, cloudy images, even when relatively close in time, are largely ignored by the model. 

Taken together, these two experiments illustrate how the cross-attention mechanism redistributes its weights depending on the temporal configuration and the availability of optical information. When a large temporal gap separates the target date from usable optical observations (either because no acquisition is available or because existing images are contaminated by clouds) the model assigns significantly higher attention to the temporally closest radar acquisition to guide the reconstruction. In the opposite case, when cloud-free optical data are available near the target date, the model concentrates its attention on these observations and assigns negligible weight to radar inputs. This behavior is intuitive, as optical imagery directly provides the spectral information to be reconstructed. However, it may also represent a potential limitation: in situations involving very rapid surface changes that are more readily detectable in radar signals, an under utilization of SAR data could reduce the model’s sensitivity to such abrupt events.

\subsection{Temporal encoding}

We conducted an ablation study to evaluate the impact of the temporal encoding strategy. In the proposed configuration, input dates are encoded relative to the target date, meaning the model receives the temporal offset ($\Delta{d}= d_{i} - d_{traget}$) between each observation and the prediction timestamp. This explicitly informs the network about temporal proximity.

We trained a second model using a standard encoding scheme, where each input date is represented independently (e.g., via sinusoidal day-of-year encoding) without computing the delta to the target date. In this case, the model must implicitly infer temporal relationships. Both models were trained under identical settings. This comparison isolates the effect of target-relative encoding and allows us to assess whether explicitly modeling temporal distance improves reconstruction performance.

Tab. ~\ref{temproal_encoding} shows that the target-relative ($\Delta{d}$) temporal encoding consistently improves performance compared to absolute date encoding. Lower MAE and RMSE, along with higher PSNR, are obtained when the model explicitly receives the temporal offset to the target date. These results suggest that providing relative temporal information helps the network better exploit temporal proximity, whereas absolute encoding requires the model to infer these relationships implicitly, leading to slightly weaker performance.

\begin{table}[h]
\centering
\caption{Effect of target-conditioned ($\Delta{d}$) temporal encoding compared to absolute date encoding.}
\begin{tabular}{llccc}
\hline
Model &  MAE & RMSE & PSNR \\
\hline
Delta &  \textbf{\fpeval{round((  0.015 + 0.019) / 2, 3)}} & \textbf{\fpeval{round(( 0.025 +  0.031) / 2, 3)}} & \textbf{\fpeval{round(( 32.557 + 30.536) / 2, 3)}} \\
\hline
Without delta & \fpeval{round(( 0.021 + 0.017 ) / 2, 3)} & \fpeval{round(( 0.034 + 0.029 ) / 2, 3)} & \fpeval{round(( 29.821 + 31.443) / 2, 3)} \\
\hline
\end{tabular}
\label{temproal_encoding}
\end{table}

\section{Limitations and future work}

Despite its overall performance, the proposed model presents several limitations. It tends to underestimate very high reflectance values, such as reflections over artificial surfaces or exceptionally bright crops. These extreme radiometric conditions are rare in the training set and are therefore implicitly regularized by the network, resulting in a smoothing effect and reduced accuracy for peak values.

\begin{figure}[htbp]
\centering
\includegraphics[width=0.4\textwidth]{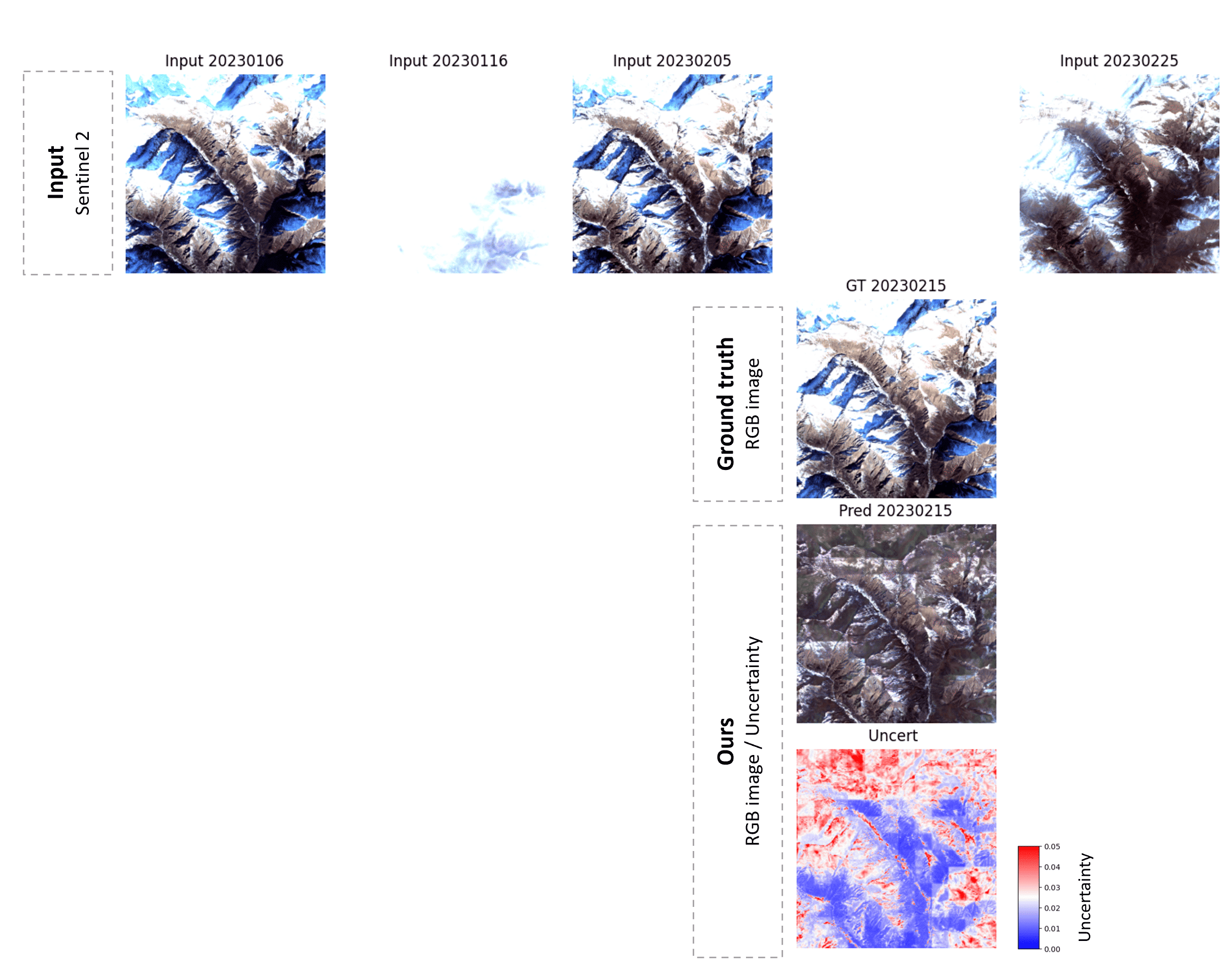}
\caption{Qualitative illustration of model limitations in snowy mountainous regions: confusion between snow and clouds leads the network to reconstruct the presumed underlying land surface beneath snow-covered areas, resulting in physically inconsistent predictions and elevated uncertainty.}
\label{limitation_neige}
\end{figure}

Another limitation is observed in mountainous regions affected by snow. The model has never been trained on mountain environments and has not been exposed to snow-covered scenes. As a result, it frequently confuses snow with clouds. Since bright cloud-contaminated pixels are treated as partially obscured observations during training, the model applies the same strategy to snow-covered areas: instead of preserving the snow signal, it attempts to reconstruct what it assumes lies beneath (e.g., bare soil or vegetation), as it would for cloud-covered regions, as shown in Fig.~\ref{limitation_neige}.

In addition, strong patching artifacts can be observed over such scenes, with visible discontinuities between adjacent tiles. These spatial inconsistencies further highlight the limited robustness of the model when confronted with out-of-distribution conditions.

Importantly, these problematic regions are generally associated with elevated predictive uncertainty. Although the reconstructed reflectance is physically incorrect, the increased uncertainty indicates that the model is not confident in its predictions, providing a useful indicator that results in these areas should be interpreted with caution.

Improving robustness would require extending the training dataset to include mountainous and snow-covered landscapes, and potentially incorporating additional spectral bands, to better discriminate between snow, clouds, and other highly reflective surfaces.

\section{Conclusion}

This work introduced a probabilistic deep learning framework for the densification and forecasting of Sentinel-2 time series from sparse and irregular multimodal observations. By formulating the task as target-conditioned image generation, the proposed model naturally supports both temporal interpolation and extrapolation within a unified architecture.

In interpolation settings, the model achieves strong and consistent performance across land-cover types, including highly dynamic cropland areas characterized by pronounced seasonal variations. The integration of Sentinel-1 SAR data provides complementary structural information, particularly in situations where optical observations are temporally distant or affected by cloud contamination. Cross-attention analyses show that the model adaptively redistributes its focus depending on data availability, leveraging radar information when optical inputs are limited and relying primarily on temporally close usable optical acquisitions when available.

In extrapolation mode, although the task is inherently more challenging, the model preserves coherent radiometric trends and captures the main seasonal vegetation dynamics. A key contribution of this work lies in the explicit modeling of predictive uncertainty. The Laplace-based probabilistic formulation produces well-calibrated uncertainty estimates that increase when input data are sparse, temporally distant, or less informative, providing a meaningful indicator of prediction reliability.

Overall, the proposed framework highlights the potential of uncertainty-aware multimodal spatio-temporal modeling for generating continuous and reliable optical satellite time series, opening perspectives for applications in long-term monitoring and decision support.

\end{document}